%% file: main.tex
\documentclass{article}
\usepackage[utf8]{inputenc}
\usepackage[a4paper, left=1in, right=1in, top=1in, bottom=1in]{geometry}
\usepackage{authblk}

\usepackage[sort&compress,numbers]{natbib}
\usepackage{mathpazo}
\usepackage{microtype}
\usepackage{graphicx}
\usepackage{booktabs}
\usepackage{hyperref}
\usepackage{amsmath}
\usepackage{amssymb}
\usepackage{mathtools}
\usepackage{amsthm}
\usepackage{mathrsfs}
\usepackage{thmtools}
\usepackage{url}
\usepackage{xspace}
\usepackage[parfill]{parskip}
\usepackage{lipsum}
\usepackage{fancyhdr}
\usepackage{enumitem}
\usepackage{algorithm}
\usepackage{algpseudocode}
\usepackage{multirow}
\usepackage{subcaption}
\usepackage[many]{tcolorbox} 
\usepackage{xcolor}    

\fancypagestyle{firststyle}
{
   \fancyhf{}
   \fancyfoot[R]{\thepage\quad}
}

\theoremstyle{plain}

\theoremstyle{definition}

\theoremstyle{remark}

\makeatletter
\DeclareRobustCommand\onedot{\futurelet\@let@token\@onedot}
\def\@onedot{\ifx\@let@token.\else.\null\fi\xspace}

\makeatother

\definecolor{grass}{RGB}{127, 186, 0}
\definecolor{grey}{HTML}{737373}
\definecolor{brick}{HTML}{F25022}

\hypersetup{
    colorlinks,
    linkcolor={grass!50!black},
    citecolor={blue!50!black},
    urlcolor={grass!70!black}
}

\newtcolorbox{titlebox}[2][]{
    arc=3mm,
    lower separated=false,
    colback=white,
    colframe=grey,
    coltitle=grass!85!black,
    fonttitle=\Large\bfseries,
    boxed title style={
        size=small,
        colback=white,
        colframe=white,
    },
    enhanced,
    attach boxed title to top right={xshift=3mm, yshift=1mm-\tcboxedtitleheight},
    adjusted title=#2
}

\makeatletter
\renewcommand\AB@affilsepx{, \protect\Affilfont}

\makeatother

\makeatletter
\renewcommand{\maketitle}{
    \begin{flushleft}
    \vskip 1em
    {\bfseries \huge \@title \par}
    \vskip 0.5em
    {\bfseries \small \@author \par}
    \vskip 0.6em
    {\footnotesize
    \begin{tabular}{@{}r@{\hspace{0.45em}}l@{}}
    \textsuperscript{*} & Equal contribution \\
    \textsuperscript{\ensuremath{\dagger}} & Work done at Microsoft Research Asia
    \end{tabular}
    \par}
    \vskip 2em
    \end{flushleft}
}
\makeatother

\title{\LARGE Beyond Monotonic Progress: \\ {\LARGE Retry-Supervised Value Learning for Robot Imitation}}

\author[1]{Xinyao Qin\textsuperscript{*\ensuremath{\dagger}}}
\author[2]{Junjie Lu\textsuperscript{*\ensuremath{\dagger}}}
\author[3]{Kaixin Wang}
\author[3]{Chuheng Zhang}
\author[4]{Sinjae Kang}
\author[4,5]{Kimin Lee}
\author[2]{Min Xu}
\author[1]{\\ Bin Liang}
\author[1]{Jun Yang}
\author[3]{Li Zhao}

\affil[1]{Tsinghua University}
\affil[2]{University of Technology Sydney}
\affil[3]{Microsoft Research Asia}
\affil[4]{KAIST}
\affil[5]{Config}

\date{}
\newcommand{\method}{ReTVL}

\begin{document}

\thispagestyle{firststyle}

\begin{titlebox}{\smash{\raisebox{1pt}{\includegraphics[width=1cm]{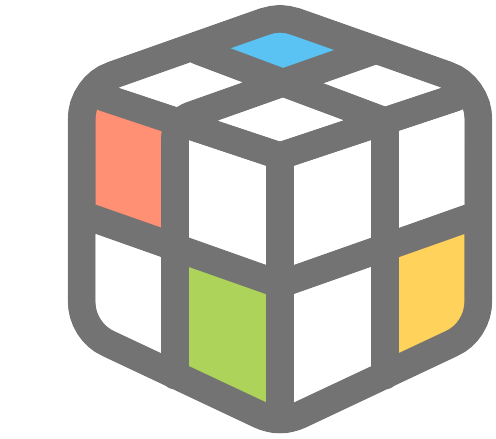}}}}
    \maketitle
    Human demonstrations for robot imitation learning often contain mistakes and corrective behaviors, such as imprecise grasps, object misalignment, unstable contact, and repeated attempts. While these segments are commonly treated as noisy or suboptimal data, they provide valuable evidence about when execution deviates from a desirable path and how task feasibility can be restored. However, existing reward and value models often rely on monotonic progress assumptions, which capture coarse task advancement but may overlook local execution errors and corrective behaviors in imperfect demonstrations. In this work, we propose \method{} (\textbf{ReT}ry-Supervised \textbf{V}alue \textbf{L}earning), a framework for learning mistake-sensitive value functions from mixed-quality robot demonstrations by leveraging retry events as sparse supervision. \method{} captures the local degradation-and-recovery structure around mistakes by combining global progress calibration with local pairwise preference learning induced by sparsely annotated retry keypoints. The learned value model is then used to reweight demonstration chunks for downstream behavior cloning, reducing the influence of harmful execution errors while preserving useful corrective behaviors. Experiments on real-robot manipulation tasks show that \method{} produces more fine-grained value estimates than progress-based baselines and improves imitation learning from imperfect demonstrations.
    \vskip 0.5em
    
    \hfill
    \begin{minipage}{0.45\linewidth}
    demo: \url{https://youtu.be/6aF6QrPg2To}
    \end{minipage}
\end{titlebox}

\pagestyle{fancy}
\fancyhf{}
\fancyhead[C]{Retry-Supervised Value Learning for Robot Imitation}
\renewcommand{\footrulewidth}{0.1pt}
\fancyfoot[R]{\thepage\quad}

\input{sec/intro}
\input{sec/related}
\input{sec/method}
\input{sec/exp}
\input{sec/conclusion}

\bibliography{main}
\bibliographystyle{main}

\newpage
\appendix

\input{app/A}

\input{app/B}

\input{app/C}
\input{app/D}


\end{document}

%% file: sec/intro.tex
\section{Introduction}
Vision-Language-Action (VLA) models have significantly improved the capabilities of general-purpose robotic policies in recent years~\citep{brohan2023rt1roboticstransformerrealworld,brohan2023rt2visionlanguageactionmodelstransfer,black2026pi0visionlanguageactionflowmodel,intelligence2025pi05visionlanguageactionmodelopenworld,amin2025pistar,kim2024openvlaopensourcevisionlanguageactionmodel, liu2024rdt, dexvla, bu2025agibot}.
By integrating visual and language understanding into action generation, these models have demonstrated great generality across diverse manipulation tasks~\citep{jiang2023vimageneralrobotmanipulation,octomodelteam2024octoopensourcegeneralistrobot,zhao2025vlasvisionlanguageactionmodelspeech,chen2025villaxenhancinglatentaction,shukor2025smolvlavisionlanguageactionmodelaffordable,li2025bridgevlainputoutputalignmentefficient}.
At the same time, applying VLAs to downstream real-world tasks still relies heavily on human demonstration data, which can be noisy and imperfect in practice~\citep{laskey2017dartnoiseinjectionrobust,wu2019imitationlearningimperfectdemonstration,zhang2021confidence,kelly2019hgdagger}.
Such data often contains hesitation, imprecise manipulations, and repeated attempts, yet these noisy or suboptimal segments are often ignored or simply discarded.
In this work, we study the problem of value learning from such imperfect demonstration data.

\begin{figure}[t]
  \centering
  \includegraphics[width=0.99\linewidth,]{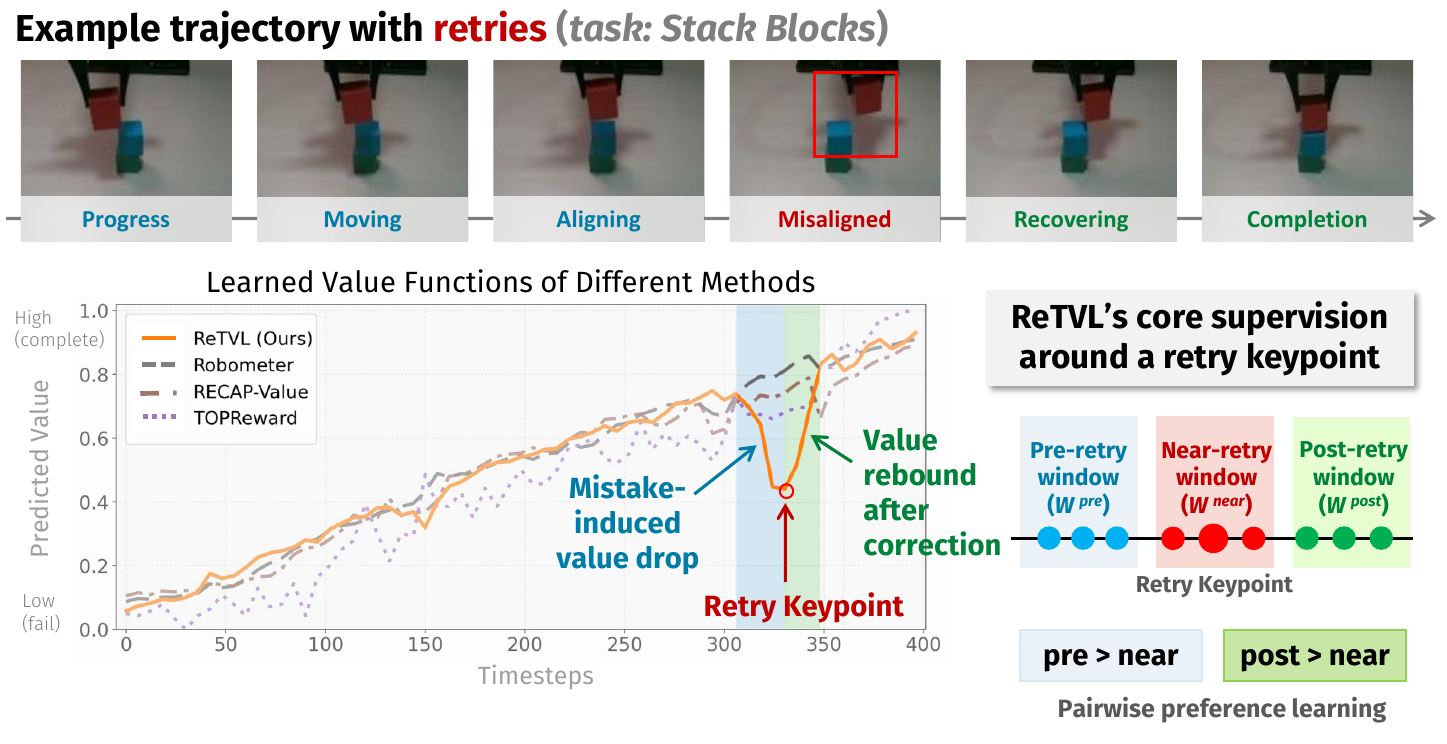}
  \caption{\textbf{\method{} turns retry events into pairwise value supervision.} Progress-based value models may overlook subtle execution errors and assign overly smooth increasing values. \method{} uses retry keypoints to learn local value drops before correction and rebounds after recovery, enabling better identification of harmful and corrective trajectory segments.}
  \label{fig:head}
\end{figure}

Recent work learns value or reward functions from robot data by estimating task progress~\citep{lee2026roboreward,liang2026robometer,chen2026topreward,amin2025pistar, chen2025sarm, mao2026arm, tan2025robo, yang2026rise, li2025gr}.
These methods typically assume that task progress can be represented as a monotonically increasing scalar signal along a trajectory, which fails to capture intermediate imperfect segments such as mistakes and retries.
For example, as illustrated at the top of Figure~\ref{fig:head}, a teleoperator may overshoot or make mistakes and later correct their behavior.
Modeling such trajectories as monotonic progress can therefore lead to inaccurate value functions, as shown by the line plot in Figure~\ref{fig:head}.
Consequently, using these inaccurate value estimates for downstream robot imitation learning can degrade the performance of the resulting policy.

Addressing this issue requires explicitly accounting for retry events during value learning.
Such retry events are common in robot demonstration data and often incur little additional labeling cost, yet they are typically ignored or discarded.
An intuitive observation is that a retry event often indicates that execution has deviated from a desirable path and is subsequently brought back through corrective behavior.
Rather than treating such retries as noisy or low-quality data, we leverage them as sparse supervisory signals for value learning.
Specifically, the temporal neighborhood around a retry reveals a local value structure: value should decrease as execution drifts toward a degraded state that requires correction, and increase as corrective behavior restores task feasibility.
These retry events therefore induce local pairwise preferences, allowing the value model to capture value drops and rebounds around subtle execution errors without requiring dense frame-level progress labels.


Based on this insight, we propose \method{}, a new value learning framework that leverages retry events as sparse supervision.
Specifically, \method{} constructs state pairs around retry events and introduces a preference-based loss to train the value function.
We find that this design is critical for accurate value estimation, whereas a naive strategy that simply penalizes retry events within conventional progress-regression-based value learning performs poorly.
We evaluate \method{} on four real-world robot manipulation tasks against three representative value estimation methods: TOPReward~\citep{chen2026topreward}, Robometer~\citep{liang2026robometer}, and RECAP-Value~\citep{amin2025pistar}.
The results show that \method{} accurately assigns lower values to harmful segments and higher values to useful corrective behaviors, capturing the value dynamics around retry events without collapsing into an uninformative monotonic trend.
We further apply the learned value functions to downstream policy learning by reweighting action chunks from suboptimal demonstration datasets.
Compared with standard behavior cloning and progress-based baselines, \method{} achieves higher success rates across diverse manipulation tasks.
Our contributions are summarized as follows:


\begin{enumerate}[label=(\arabic*), leftmargin=*, topsep=0pt]
\item We identify retry events as informative supervision for value learning, revealing local value drops and rebounds that are often hidden by coarse progress labels.
\item We propose \method{}, which combines global progress supervision with local relative preferences to learn values that reflect both execution errors and corrective behaviors.
\item We show that \method{} improves both value estimation and downstream policy learning on mixed-quality demonstrations, outperforming progress-based value models and their corresponding value-weighted policy learning baselines.
\end{enumerate}

%% file: sec/related.tex
\section{Related Works}
\subsection{Learned Value and Reward Models for Robot Learning}

Learned reward and value models have been widely used to provide dense
supervision for robot learning when task success is sparse or difficult to
specify manually. Early work infers rewards from demonstrations through inverse
reinforcement learning, but often faces scalability challenges in
high-dimensional, long-horizon problems~\citep{ng2000algorithms,abbeel2004apprenticeship,ziebart2008maximum,finn2016guided}. 
Other work learns dense reward signals from human videos~\citep{chen2021dvd,ma2022vip},
cross-embodiment videos~\citep{alakuijala2024vlc}, or language-image
supervision~\citep{ma2023liv,alakuijala2024vlc}, enabling reward estimation for
downstream manipulation tasks.
More recent methods scale reward modeling with large robotics datasets and
vision-language models, learning general-purpose reward or value models from
progress labels~\citep{lee2026roboreward, amin2025pistar}, success signals~\citep{liang2026robometer, li2025gr}, trajectory comparisons~\citep{liang2026robometer}, or stage
annotations~\citep{chen2025sarm,tan2025robo}. 
These learned values have been used for success-failure prediction, data
filtering, and policy improvement in downstream tasks~\citep{chen2025sarm,amin2025pistar}. 
While prior methods mainly capture global progress and success likelihood, they may fail to reflect subtle execution errors. Our method instead leverages retry-based supervision to learn local mistake-and-recovery dynamics.

\subsection{Learning Policy from Suboptimal Demonstrations}
Learning from suboptimal demonstrations has been widely studied to reduce the need for clean expert data. Prior imitation learning methods estimate demonstration quality through
confidence scores~\citep{wu2019imperfect}, learned reliability or confidence
estimation~\citep{zhang2021confidence}, distribution matching~\citep{kim2022demodice},
or discriminator-based weighting~\citep{xu2022dwbc}, and use these signals to
select or reweight useful demonstrations for imitation learning.
Other work uses ranked demonstrations~\citep{brown2019trex,brown2020drex} or
preference feedback~\citep{christiano2017preferences,sadigh2017active,palan2019dempref}
to guide policy optimization beyond suboptimal demonstrators without manually
specified rewards.
A more closely related direction learns value functions and applies offline RL
or weighted behavior cloning to prioritize useful actions during policy
learning~\citep{amin2025pistar,chen2025sarm,yang2026aloe,huang2020awr}. Our method follows this weighted BC paradigm, but derives the weights from a
retry-supervised value model, which more directly suppresses local mistake
segments in mixed-quality demonstrations.

%% file: sec/method.tex
\section{Method}
\subsection{Problem Setup}
\paragraph{Value Function.}
Let \(h_t=\mathbf{o}_{t-H+1:t}\) denote the history observation window of size $H$ ending at time \(t\). 
Given \(h_t\) and a task instruction \(\ell\), the value model estimates the current task progress as \(V_\theta(h_t,\ell)\in[0,1]\). 
In practice, we use a distributional value function that predicts a categorical distribution over \(K\) discretized progress bins for the stability and reliability of value estimation. 
Let \(p_\theta(k\mid h_t,\ell)\) be the predicted probability of the \(k\)-th bin, and \(b_k\) denote the center value of that bin. 
We compute the scalar value as the expected progress under the predicted distribution:
\begin{equation}
    V_\theta(h_t,\ell)
    =
    \sum_{k=0}^{K-1} b_k\,p_\theta(k \mid h_t,\ell).
\end{equation}

\paragraph{Progress Supervision.}
To provide a global progress reference, we apply absolute progress supervision to frames outside retry neighborhoods. For a successful trajectory of length \(T\), the target progress at timestep \(t\) is defined as
\(v_t^\star = t/T\in[0,1]\), which provides a coarse estimate of how far the execution has advanced. For failed trajectories, we assign \(v_t^\star=0\) at terminal frames.
Each target progress value \(v^\star\) is discretized into a target bin \(b^\star\), and the model is trained to predict this bin with a cross-entropy loss~\citep{amin2025pistar}:
\begin{equation}
    \mathcal{L}_{\mathrm{abs}}
    =
    -
    \mathbb{E}_{(h_t,b_t^\star)}
    \log p_\theta(b_t^\star\mid h_t,\ell),
    \label{eq:abs_loss}
\end{equation}
This objective provides a global progress signal by assigning lower values to earlier states and higher values to later states along the trajectory.

\paragraph{Data Annotations.}
We consider a demonstration dataset \(\mathcal{D}=\{\tau_i\}_{i=1}^{N}\), 
where each trajectory \(\tau_i\) is associated with a binary outcome label 
\(y_i\in\{0,1\}\) indicating task failure or success. 
For trajectories that contain retry behaviors, we additionally annotate a set 
of retry keypoints \(\mathcal{R}_i=\{r_{i,j}\}_{j=1}^{M_i}\). 
Each keypoint \(r_{i,j}\) marks the start of the \(j\)-th corrective behavior 
in trajectory \(\tau_i\). 
Notably, we do not annotate the full retry segment such as the start of deviation or the completion of recovery. Instead, we only annotate the moment when corrective behavior starts. This choice makes the supervision more reliable and easier to obtain, since the full retry segment can be annotator-dependent, while the start of correction is usually easy to label during data collection or extracted from existing human-correction datasets~\citep{kelly2019hgdagger,xu2026compliant}.

\begin{figure}[t]
  \centering
  \includegraphics[width=0.99\linewidth,]{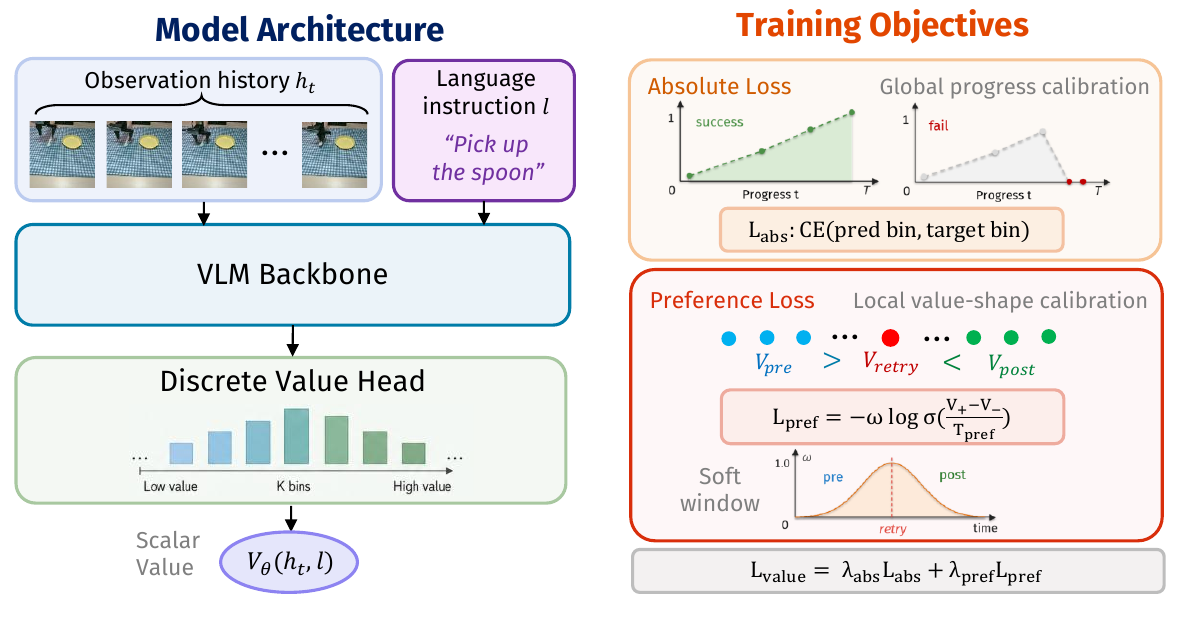}
  \caption{
\textbf{\method{} learns retry-sensitive value estimates from sparse retry annotations.}
The model takes an observation history and language instruction as input, and predicts a scalar value through a VLM backbone and discrete value head.
Training combines absolute progress calibration with retry-induced preference supervision, where values drop near retry states and rebound after recovery.
}
  \label{fig:frame}
\end{figure}

\subsection{Pairwise Value Learning from Retry Events}
\label{sec:pair-window}
\paragraph{Pair Construction.}
For each retry keypoint \(r_{i,j}\), we construct preference pairs based on a simple but natural drop-and-rebound assumption: values should decrease before the correction starts and increase after effective corrective behavior begins.
To this end, we divide the retry neighborhood into three regions:
\begin{align}
    \mathcal{W}^{\mathrm{pre}}_{i,j}
    &=
    \left\{
    t \mid r_{i,j}-\Delta_{\mathrm{pre}} \le t
    < r_{i,j}-\Delta_{\mathrm{near}}
    \right\},
    \\
    \mathcal{W}^{\mathrm{near}}_{i,j}
    &=
    \left\{
    t \mid r_{i,j}-\Delta_{\mathrm{near}} \le t
    \le r_{i,j}+\Delta_{\mathrm{near}}
    \right\},
    \\
    \mathcal{W}^{\mathrm{post}}_{i,j}
    &=
    \left\{
    t \mid r_{i,j}+\Delta_{\mathrm{near}} < t
    \le r_{i,j}+\Delta_{\mathrm{post}}
    \right\}.
    \label{eq:retry_windows}
\end{align}
Here, \(\Delta_{\mathrm{near}}\) defines a small neighborhood around the retry keypoint, while \(\Delta_{\mathrm{pre}}\) and \(\Delta_{\mathrm{post}}\) define the temporal ranges used for sampling before and after the retry, respectively.
We then sample preference pairs \((h^+,h^-)\) such that the retry neighborhood forms a local value minimum, as it corresponds to a degraded state where correction becomes necessary. 
On the pre-retry side, values are expected to be higher than those in the near-retry region and gradually decrease before the retry keypoint. On the post-retry side, values should increase over time and remain higher than those in the near-retry region.
This construction provides local supervision for both the value drop before a retry and the value rebound after corrective behavior.

\paragraph{Soft-Window Weighting.}
The retry neighborhoods defined above are only approximate partitions of the local retry region and can be ambiguous in practice.
Assigning equally strong preference constraints to all sampled pairs may therefore introduce noisy
supervision, especially for pairs that rely on frames far from the retry
keypoint or close to uncertain window boundaries. To mitigate this issue, we use a soft-window weighting scheme.
For each sampled pair \((h^+,h^-)\) around retry keypoint \(r_{i,j}\), we compute the weight using the endpoint farther from the retry keypoint, which corresponds to the preferred history \(h^+\) under the drop-and-rebound assumption. Let \(t^+\) denote the ending timestep of \(h^+\). We define
its temporal distance to the retry keypoint as
\begin{equation}
    d(h^+; r_{i,j}) = |t^+ - r_{i,j}|.
\end{equation}
The soft weight is then defined as
\begin{equation}
    w(h^+,h^-)
    =
    \exp\!\left(-d(h^+;r_{i,j})/\tau_w\right),
    \label{eq:soft_weight}
\end{equation}
where \(\tau_w\) controls the decay rate. This weighting assigns stronger
supervision to pairs whose anchor is closer to the retry keypoint and
smoothly downweights pairs whose anchor is farther away, making the preference
loss less sensitive to imprecise retry-window boundaries.

\paragraph{Preference Loss.}
Given a sampled preference pair \((h^+,h^-)\), where \(h^+\) should have a
higher value than \(h^-\), we optimize the soft-weighted pairwise logistic loss~\citep{bradley1952rank,christiano2017deep}:
\begin{equation}
    \mathcal{L}_{\mathrm{pref}}
    =
    -
    \mathbb{E}_{(h^+,h^-)}
    w(h^+,h^-)
    \log \sigma
    \left(
    \frac{
    V_\theta(h^+,\ell)-V_\theta(h^-,\ell)
    }{T_\mathrm{pref}}
    \right).
    \label{eq:pref_loss}
\end{equation}
Here, \(T_\mathrm{pref}\) is a temperature parameter that controls the sharpness of the preference comparison, \(\sigma(\cdot)\) denotes the sigmoid function, and \(w(h^+,h^-)\) is the soft-window weight. This loss enforces local value
ordering around retry events, encouraging value drops before correction and
value recovery after corrective behavior.

We then train the value model on a mixed-quality demonstration dataset combining absolute loss and preference loss. Absolute progress samples are drawn from
all trajectories, except for frames inside retry neighborhoods where
monotonic progress labels may be unreliable. In contrast, retry-based
preference pairs are sampled from retry regions to provide
local relative supervision. The two supervision sources are mixed with a fixed
sampling ratio \(\rho_{\mathrm{pref}}\).
The total training objective combines global progress calibration and local
retry-based preference learning, balanced by \(\lambda_{\mathrm{abs}}\) and
\(\lambda_{\mathrm{pref}}\):
\begin{equation}
    \mathcal{L}_{\mathrm{value}}
    =
    \lambda_{\mathrm{abs}}\mathcal{L}_{\mathrm{abs}}
    +
    \lambda_{\mathrm{pref}}\mathcal{L}_{\mathrm{pref}}.
    \label{eq:value_objective}
\end{equation}

\subsection{Value-Guided Behavior Cloning}
\label{sec:policy-weighting}
We use our value model to guide policy learning from mixed-quality demonstrations
to evaluate whether it brings practical improvements to downstream robot learning.
This evaluation follows the weighted behavior cloning procedure from prior
work~\cite{chen2025sarm,peng2019advantageweightedregressionsimplescalable,nair2021awacacceleratingonlinereinforcement}.
For an action chunk starting at time \(t\) with stride \(\Delta_a\), we compute
its weight from the predicted progress improvement and normalize the weight using
offline dataset statistics:
\begin{equation}
r_t
=
V_\theta(h_{t+\Delta_a},\ell)-V_\theta(h_t,\ell),
\qquad
\alpha_t
=
\mathrm{clip}
\left(
\frac{
r_t - (\mu - 2\sigma)
}{
4\sigma + \epsilon
},
0, 1
\right),
\label{eq:chunk_weight}
\end{equation}
where \(\mu\) and \(\sigma\) are the offline mean and standard deviation of
value improvements over the training dataset.
The policy \(\pi_\psi\) is trained with a normalized weighted BC objective, which downweights chunks with low or negative predicted progress improvement while emphasizing chunks that move toward higher predicted task progress:
\begin{equation}
    \mathcal{L}_{\mathrm{wBC}}(\psi)
    =
    \frac{
    \sum_t \alpha_t\,
    \ell_{\mathrm{BC}}\!\left(\pi_\psi(h_t), a_t\right)
    }{
    \sum_t \alpha_t + \epsilon
    }.
    \label{eq:wbc}
\end{equation}

%% file: sec/exp.tex
\section{Experiments}
\label{sec:experiments}
Our experiments aim to study whether \method{} can produce mistake-sensitive value estimates for subtle execution errors and thereby improve downstream policy learning. Specifically, we organize our experiments to answer the following questions:

\noindent
\textbf{(Q1) Value Evaluation:} Does our value model better capture sublte
execution errors and show value changes than progress-based reward or
value models?

\noindent
\textbf{(Q2) Policy Learning:} Given the same mixed-quality demonstration dataset,
can our learned values identify more useful trajectory segments and improve
downstream imitation learning?

\noindent
\textbf{(Q3) Ablation and Analysis:} Which components of \method{} are most important, and why does it improve value quality and policy performance?

\paragraph{Baselines.}
We compare \method{} against the strongest available reward and value models
that can provide dense scores for robot learning. These baselines cover three representative forms of dense supervision for robot learning: zero-shot VLM-based progress estimation, general-purpose robotic reward modeling, and progress-supervised value learning.

\begin{itemize}[leftmargin=*,labelsep=0.5em]
    \item \textbf{TOPReward~\citep{chen2026topreward}} directly prompts a pre-trained VLM to judge whether
    the task is completed and use the normalized probability of the ``true'' token as a progress score.

    \item \textbf{Robometer~\citep{liang2026robometer}} is a general-purpose robotic reward model trained on large-scale robot data. It uses trajectory rewinding and trajectory-level preferences to learn rewards for failed and suboptimal executions beyond progress supervision.

    \item \textbf{RECAP-Value} denotes our implementation of the value component in RECAP~\citep{amin2025pistar}. It uses Monte Carlo step-to-success targets to learn a progress value function with a discrete value head.
\end{itemize}

\paragraph{Tasks and Datasets.}
We evaluate our method on 4 real-robot manipulation tasks using a Piper robot
arm, as illustrated in Appendix A.
\textit{Pick up Spoon} is a
relatively simple pick-and-place task, while the other three tasks are
long-horizon tasks covering different manipulation challenges:
\textit{Stack Blocks} requires fine-grained alignment when stacking three blocks,
\textit{Fold Towel} involves deformable-object manipulation, and
\textit{Open Drawer} requires interacting with an articulated object.
For each task, we use 30 annotated trajectories to train the value model and evaluate the learned values on 20 held-out trajectories.
The trained value model is then used to score a separate mixed-quality dataset for downstream policy learning, consisting of 200 trajectories per task except \textit{Pick up Spoon}, which uses 80 trajectories.

\subsection{Value Evaluation}
\label{sec:value_eval}
We first evaluate whether \method{} can achieve strong global value estimation while better capturing subtle execution errors than baselines.
We report both global and local metrics in Table~\ref{tab:value_eval}.
The global metrics include Value-Order Correlation (VOC) on clean successful trajectories and Success-Fail
Detection (S/F Det.).
They evaluate whether the value model captures temporal progress in successful executions and distinguishes successful trajectories from failed ones, reflecting its basic ability to estimate global task progress and terminal success.
The local metrics evaluate whether the value model produces a localized value drop around subtle execution errors and recovers after correction.
Drop PR-AUC and Drop Probability measure the alignment and frequency of such drops near retry events, while Pre $>$ Retry and Post $>$ Retry measure whether pre-mistake and post-recovery states are valued higher than retry states.
We provide formal definitions in Appendix B.

\begin{table}[t]
\centering
\small
\setlength{\tabcolsep}{4.2pt}
\renewcommand{\arraystretch}{1.08}
\caption{
Value evaluation across four real-world tasks. Global metrics measure trajectory-level success calibration, while local metrics measure retry-related mistake sensitivity.
}
\vspace{0.2cm}
\label{tab:value_eval}
\begin{tabular}{lcc|cccc}
\toprule
\multirow{2}{*}{Method}
& \multicolumn{2}{c|}{Global Metrics}
& \multicolumn{4}{c}{Local Metrics} \\
\cmidrule(lr){2-3} \cmidrule(lr){4-7}
& VOC $\uparrow$
& S/F Det. $\uparrow$
& Drop AUC $\uparrow$
& Drop Prob. $\uparrow$
& Pre $>$ Retry $\uparrow$
& Post $>$ Retry $\uparrow$ \\
\midrule
TOPReward
& 0.903 & 0.844 & 0.264 & 0.612 & 0.329 & 0.917 \\
Robometer  
& \textbf{0.994} & 0.964 & 0.237 & 0.435 & 0.086 & 0.828 \\
RECAP-Value      
& 0.985 & \textbf{1.000} & 0.372 & 0.721 & 0.296 & 0.854 \\
\method{} 
& 0.987 & \textbf{1.000} & \textbf{0.797} & \textbf{0.874} & \textbf{0.740} & \textbf{0.967} \\

\bottomrule
\end{tabular}
\end{table}

\begin{figure}[t]
  \centering
  \includegraphics[width=1.0\linewidth,]{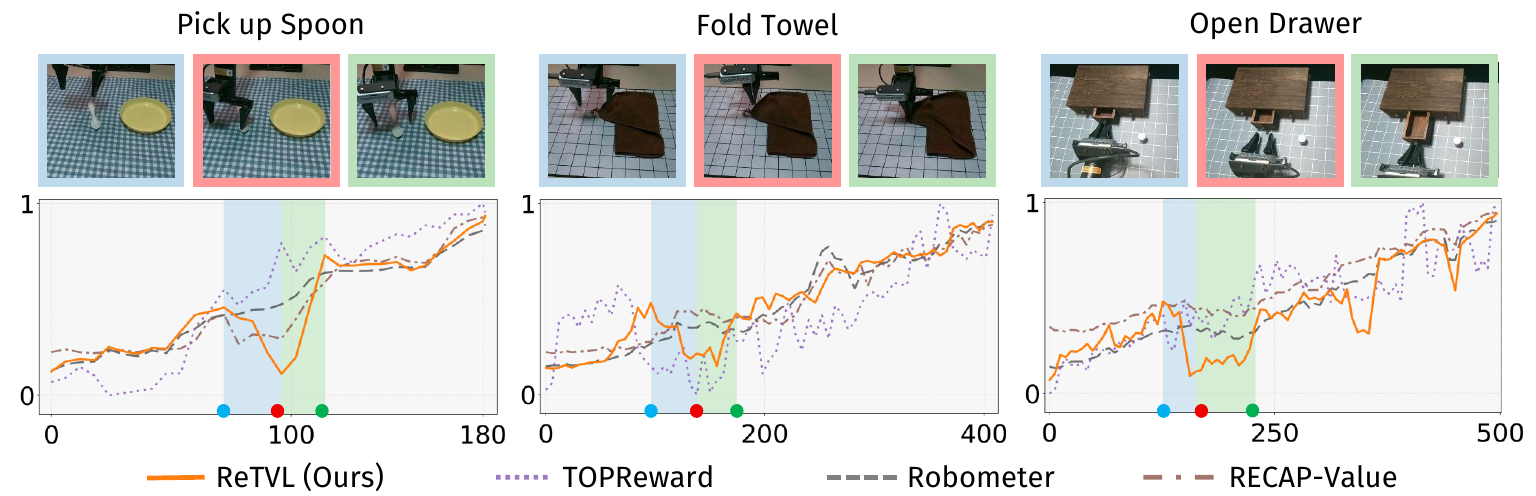}
  \caption{\textbf{Visualization of value evaluation.} We show value predictions on three other tasks beyond stack blocks. \method{} captures local value drops around retry keypoints and rebounds after correction more clearly than progress-based baselines.}
  \label{fig:visualization}
\end{figure}

Table~\ref{tab:value_eval} shows that \method{} achieves competitive global value estimation while substantially improving local mistake sensitivity.
On global metrics, \method{} obtains a VOC score of 0.987, close to the best score from Robometer, and achieves the best Success-Fail Detection score of 1.000, matching RECAP-Value. 
These results indicate that \method{} retains the basic ability to model trajectory-level progress and distinguish successful executions from failures.
The main advantage of \method{} lies in local retry-centered metrics.
It achieves consistent improvements across all four local metrics, indicating that it more reliably captures retry-related local value changes.
The improvement is especially large on Pre $>$ Retry, where \method{} reaches 0.740 while all baselines remain below 0.330.
This suggests that existing methods often fail to assign lower values to bad actions or problematic states that trigger retries, as illustrated in Figure~\ref{fig:visualization}.
Notably, although Robometer achieves the highest VOC score, its Pre $>$ Retry score is the lowest, only 0.086, suggesting that strong global progress correlation does not necessarily imply sensitivity to local execution mistakes.
In contrast, \method{} induces the desired local value shape around mistakes, decreasing in degraded states and recovering after correction.

\subsection{Policy Learning}
\label{sec:policy_learning}
Next, we evaluate whether the learned value model can improve downstream imitation learning from mixed-quality demonstrations.
We compare \textbf{\method{}-BC} against two baselines: \textbf{Standard BC}, which trains on all demonstrations with uniform weights, and \textbf{RECAP-BC}, which uses RECAP-Value to weight demonstration chunks.
RECAP-Value is chosen because it provides the strongest local mistake detection among the baselines while maintaining competitive global value estimation in above Section~\ref{sec:value_eval}.
For both value-weighted methods, we split trajectories into action chunks and apply the same weighting rule from Section~\ref{sec:policy-weighting}.
The only difference is whether the weights are computed by RECAP-Value or our \method{} model.

\begin{figure}[t]
\centering
\begin{minipage}[b]{0.48\linewidth}
\vspace{0pt}
\centering
\setlength{\tabcolsep}{4.5pt}
\renewcommand{\arraystretch}{1.08}
\begin{tabular}{lccc}
\toprule
Task & \shortstack{Standard\\BC} & \shortstack{RECAP\\-BC} & \shortstack{\method{}\\-BC} \\
\midrule
Pick up Spoon   & 60 & 65 & 85 \\
Stack Blocks & 45 & 80 & 95 \\
Fold Towel   & 50 & 65 & 80 \\
Open Drawer  & 10 & 40 & 60 \\
\midrule
Average      & 41 & 63 & 80 \\
\bottomrule
\end{tabular}
\vspace{0.4em}
\captionof{table}{
Success rate (\%) of the learned policies on 4 real-robot tasks, calculated from 20 trials per task.
}
\label{tab:policy_learning}
\end{minipage}
\hfill
\begin{minipage}[b]{0.48\linewidth}
\vspace{0pt}
\centering
\includegraphics[width=\linewidth]{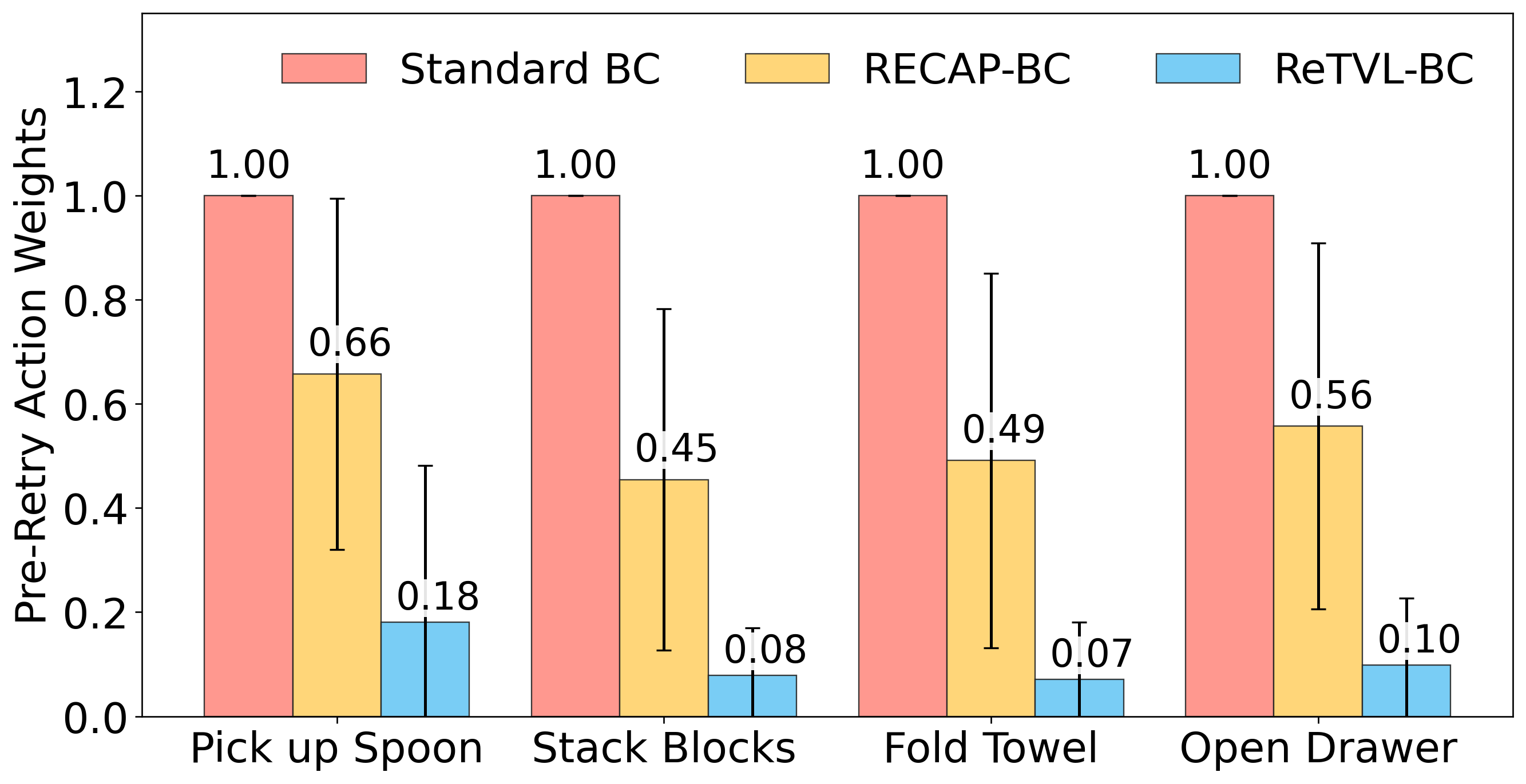}
\caption{
Average training weight assigned to annotated bad-action chunks in recovery
trajectories. Lower is better.
}
\label{fig:bad_action_weight}
\end{minipage}
\end{figure}

Table~\ref{tab:policy_learning} shows that \method{}-BC consistently improves downstream policy learning across all four real-robot tasks.
Standard BC is sensitive to the quality of mixed demonstrations, achieving an average success rate of only 41.25\%, as it imitates both clean successful actions and locally harmful actions from failed or retry trajectories. 
RECAP-BC improves the average success rate to 62.50\% by assigning larger weights to higher-progress chunks, but its progress-based value signal is still less effective in separating useful corrections from locally mistaken actions. 
In contrast, \method{}-BC achieves the best average success rate of 80.00\%, improving over Standard BC by +38.75\% and over RECAP-BC by +17.50\%.
These improvements suggest that retry-supervised values produce a better imitation-learning distribution by down-weighting mistake segments while emphasizing clean or recovery-relevant actions.

We further analyze the weights assigned to bad actions in Fig.~\ref{fig:bad_action_weight}, where bad actions are manually annotated on the held-out test set from Section~\ref{sec:value_eval}.
Standard BC assigns all actions a weight of 1.0.
RECAP-BC reduces the average bad-action weight to 0.54, while \method{}-BC further lowers it to 0.11, a 80\% reduction over RECAP-BC.
\method{} also shows much smaller variance, suggesting that it consistently suppresses harmful actions rather than only down-weighting a few obvious mistakes.

\subsection{Ablation and Analysis}
\label{sec:ablation}
We conduct ablation studies to identify which components of \method{} contribute to mistake-sensitive value estimation. 
Table~\ref{tab:ablation} first shows that the retry-induced preference loss is crucial for learning mistake-sensitive values.
The \textbf{w/o Preference Loss} variant replaces our pairwise preference objective with an intuitive alternative: directly regressing to a shaped progress target with manually injected value drops around retry windows.
Although this variant preserves strong global metrics, achieving even higher VOC (0.994) and the same Success-Fail Detection (1.000), it fails on local mistake-sensitive metrics.
Drop AUC decreases sharply from 0.797 to 0.510, Pre $>$ Retry falls from 0.740 to 0.486, and Post $>$ Retry drops from 0.967 to 0.742.
These results indicate that simply adding local drop penalties to a progress target does not reliably teach the model the relative ordering between mistaken and recovered states.
In contrast, our preference loss directly supervises local comparisons around retry events, which is essential for capturing mistake-and-recovery structure.

\begin{table}[t]
\centering
\small
\setlength{\tabcolsep}{4.0pt}
\renewcommand{\arraystretch}{1.08}
\caption{
Ablation study on held-out trajectories. We report representative global and
local metrics averaged over four tasks.
}
\vspace{0.2cm}
\label{tab:ablation}
\begin{tabular}{lcccccc}
\toprule
Method
& VOC $\uparrow$
& S/F Det. $\uparrow$
& Drop AUC $\uparrow$
& Drop Prob. $\uparrow$
& Pre $>$ Retry $\uparrow$
& Post $>$ Retry $\uparrow$ \\
\midrule
\method{}
& 0.987 & 1.000 & 0.797 & 0.874 & 0.740 & 0.967\\
w/o Preference Loss
& 0.994 & 1.000 & 0.510 & 0.836 & 0.486 & 0.742\\
w/o Soft Window
& 0.940 & 1.000 & 0.785 & 0.874 & 0.707 & 0.971\\
w/o Abs. Calibration
& 0.929 & 0.967 & 0.857 & 0.872 & 0.789 & 0.959\\
\bottomrule
\end{tabular}
\end{table}

The other two ablations reveal the complementary roles of soft windowing and
absolute calibration. The \textbf{w/o Soft Window} variant removes the soft
retry-window weighting and applies retry-related supervision uniformly. Its local
metrics remain broadly comparable to the full model, but VOC decreases from
0.987 to 0.940. This suggests that hard retry supervision can create conflicts
with the absolute progress targets, especially near the boundary between normal
progress regions and retry neighborhoods. The
\textbf{w/o Absolute Calibration} variant removes the absolute progress and
success-failure calibration losses, leaving only retry-induced preference
supervision. Although it can still learn local retry-centered ordering, its
global value structure becomes weaker, with VOC dropping from 0.987 to 0.929 and
Success-Fail Detection from 1.000 to 0.967. Overall, \method{} benefits from all
three components: preference supervision provides local mistake sensitivity,
absolute calibration anchors the global value scale, and soft windowing
stabilizes their interaction. Detailed visualizations of the ablation behaviors
are provided in Appendix B.

%% file: sec/conclusion.tex
\section{Conclusion}
\label{sec:conclusion}
We presented \method{}, a retry-supervised value learning framework for robot imitation from mixed-quality demonstrations.
By using sparsely annotated retry keypoints as local supervision anchors, \method{} combines global progress calibration with retry-induced preference learning to capture both coarse task progress and local mistake-recovery structure.
The learned values further improve downstream behavior cloning by down-weighting harmful mistake segments and emphasizing useful recovery behavior.
Real-world manipulation experiments show that \method{} improves local mistake sensitivity over progress-based baselines while maintaining competitive global value estimation, leading to better imitation learning from imperfect demonstrations.
These results suggest that corrective behaviors are not merely noisy artifacts, but useful supervision for learning mistake-aware robot policies.

\paragraph{Limitations} \method{} still has several limitations.
First, we assume that values around retry events follow a local degradation-and-recovery pattern.
Although this is more flexible than a monotonic progress assumption, it may not cover all real-world correction patterns, such as exploratory retries or unsuccessful corrections.
Second, we currently use the learned value model only for offline behavior cloning reweighting, and have not explored broader closed-loop policy improvement or online RL settings.
Third, our experiments are conducted on a limited set of real-robot tasks rather than large-scale datasets, so broader evaluation is needed to assess generalization across more tasks, robots, and demonstration styles.

%% file: app/A.tex
\section{Implementation Details}
\subsection{Value Model Training}

\paragraph{Data preprocessing.}
All value models are trained and evaluated using the same local 5 Hz data
protocol. The raw robot trajectories are recorded at 30 Hz and then downsampled
by taking every sixth frame. We use this lower frame rate because adjacent 30 Hz
frames typically show very small visual changes, which adds temporal redundancy
and can make value estimation more sensitive to frame-level noise. Since the four tasks have substantially different horizons, we do not
truncate trajectories to a fixed length after resampling. Instead, we keep all
usable 5 Hz frames and construct fixed-length inputs later during sampling.

Each trajectory used for value-model training contains the task instruction,
primary-view images, wrist-view images, the success/failure outcome, and retry
keypoints. For the value-model test set, we additionally annotate bad-action
start points and back-to-normal points, which are used only for analysis and
evaluation. In all value-model training and evaluation, the model takes only the primary
camera view as visual input.
For value-model training and evaluation, the input at endpoint $t$ is an
8-frame history window $h_t = o_{t-7:t}$.
When $t<7$, we left-pad the window by repeating the first frame. The model
predicts the value of the last frame in the window, while the preceding frames
serve as temporal context.

\paragraph{Architecture and hyperparameters.}
For a fair comparison, we use Robometer-4B~\citep{liang2026robometer} as the backbone for all trainable value models, which is
pretrained on large-scale robotic data and follows the
\texttt{Qwen/Qwen3-VL-4B-Instruct}~\citep{bai2025qwen3} architecture. We attach a two-layer MLP value
head on top of the vision-language hidden state.
For history-window inputs, frames are fed as separate images in the multi-image
input. The value head is applied only to the hidden representation of the last
frame in the window.
We fine-tune the backbone with LoRA and train the value head jointly with the
LoRA parameters. All trainable methods use the same hyperparameters, summarized
in Table~\ref{tab:appendix_all_hparams}.

For ReTVL, we instantiate the pre-retry, retry-centered, and
post-recovery regions on the 5 Hz endpoint timeline as defined in Sec.~\ref{sec:pair-window}. In each training step, the
sampler draws one of the active pair types from four pair types with equal probability: pre-vs-near,
near-vs-post, pre-vs-pre and post-vs-post. 
For pre-vs-near and near-vs-post pairs, the retry-centered window is treated as
the lower-value side of the comparison. For 
pre-vs-pre and post-vs-post, we sample two endpoints from the same side of the
retry keypoint and assign the lower value to the endpoint closer to the retry.
Thus, values are encouraged to decrease as the trajectory approaches the retry
from the pre-retry side and to increase as it moves away from the retry on the
post-recovery side. We apply the same soft-window weighted
pairwise preference loss described in Sec.~\ref{sec:pair-window}. We use the same sampling parameters for all four tasks as summarized in Table~\ref{tab:appendix_all_hparams}. Algorithm~\ref{alg:repair_training}
summarizes the overall training procedure.

\begin{algorithm}[h]
\caption{Training Procedure for \method{}}
\label{alg:repair_training}
\begin{algorithmic}[1]
\Require Mixed-quality demonstration dataset $\mathcal{D}$, retry keypoints $\{\mathcal{R}_i\}$, value model $V_\theta$
\For{each training iteration}
    \State Sample trajectories $\{\tau_i\} \sim \mathcal{D}$

    \State Sample absolute progress examples outside retry neighborhoods
    \State Compute absolute progress loss $\mathcal{L}_{\mathrm{abs}}$ using Eq.~\ref{eq:abs_loss}

    \State Sample retry keypoints $r_{i,j} \in \mathcal{R}_i$ and construct $\mathcal{W}^{\mathrm{pre}}_{i,j}$, $\mathcal{W}^{\mathrm{near}}_{i,j}$, $\mathcal{W}^{\mathrm{post}}_{i,j}$
    \State Sample preference pairs $(h^+,h^-)$ and compute weights $w(h^+,h^-)$
    \State Compute preference loss $\mathcal{L}_{\mathrm{pref}}$ using Eq.~\ref{eq:pref_loss}

    \State Update $V_\theta$ with 
    $
    \mathcal{L}_{\mathrm{value}}
    =
    \lambda_{\mathrm{abs}}\mathcal{L}_{\mathrm{abs}}
    +
    \lambda_{\mathrm{pref}}\mathcal{L}_{\mathrm{pref}}
    $
\EndFor
\end{algorithmic}
\end{algorithm}


\subsection{Downstream Weighted Behavior Cloning}
\label{app:downstream_bc_details}

For downstream policy learning, we use a flow-matching Vision-Language-Action
policy with a $\pi_0$-style backbone. The policy takes the task instruction,
RGB observations, and robot proprioception as input, and predicts an
action chunk.
Given the chunk weight $\alpha_t$ computed by Eq.~\ref{eq:chunk_weight}, we apply
value-weighted behavior cloning directly to the flow-matching objective:
\[
    \mathcal{L}_{\mathrm{wBC}}
    =
    \frac{
    \sum_t \alpha_t
    \left\|
    v_\psi(z_s, h_t, s) - u_s
    \right\|_2^2
    }{
    \sum_t \alpha_t + \epsilon
    },
\]
where $z_s$ is the interpolated noisy action chunk at flow time $s$, and $u_s$
is the corresponding target velocity. Standard BC uses the same objective with
uniform chunk weights. For ReTVL-BC and RECAP-BC, the policy architecture,
training data, and optimization settings are identical; only the value model
used to compute $\alpha_t$ differs.
We fine-tune all policy parameters and use the same downstream training
hyperparameters for all methods, as summarized in
Table~\ref{tab:appendix_all_hparams}. Following the SARM-style
weighting rule, we use task-specific thresholds $\kappa$: chunks with value
improvement larger than $\kappa$ are assigned weight $1$. For each task,
$\kappa$ is chosen to roughly correspond to the 80th percentile of positive
chunk improvements.

\begin{table*}[t]
\centering
\caption{Implementation hyperparameters used in our experiments.}
\vspace{0.2cm}
\label{tab:appendix_all_hparams}
\begin{tabular}{ll|ll}
\toprule
\multicolumn{2}{c|}{\textbf{Value Model Training}} &
\multicolumn{2}{c}{\textbf{ReTVL Pair Sampling}} \\
\midrule
Hyperparameter & Value & Hyperparameter & Value \\
\midrule
Input frame rate & 5 Hz &
Pre-retry window & $[r-12, r-2]$ \\

History window & 8 frames &
Retry-centered window & $[r-1, r+1]$ \\

MLP dims & $[512,512]$ &
Post-recovery window & $[r+2, r+12]$ \\

MLP activation & GELU &
Soft-window temperature & $\tau_w=6.0$ \\

Distributional value bins & 64 &
Preference temperature & $T_{\mathrm{pref}}=0.1$ \\

MLP dropout & 0.1 &
Abs. loss weight & $\lambda_{\mathrm{abs}}=1.0$ \\

LoRA rank & 32 &
Pref. loss weight & $\lambda_{\mathrm{pref}}=3.0$ \\

\cmidrule{3-4}
LoRA alpha & 64 &
\multicolumn{2}{c}{\textbf{Downstream Policy Training}} \\

\cmidrule{3-4}
LoRA dropout & 0.05 &
Learning rate & $5\times10^{-5}$ \\

Batch size & 64 &
Training steps & 20k \\

Learning rate & $1\times10^{-4}$ &
Batch size & 256 \\

Scheduler & Cosine &
Action chunk size & 16 \\

Training steps & 500 &
Inference flow steps & 10 \\

\bottomrule
\end{tabular}
\end{table*}

\begin{figure}[t]
  \centering
  \includegraphics[width=1.0\linewidth,]{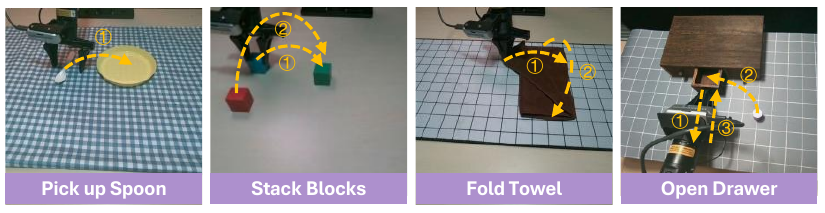}
  \caption{Real-world manipulation tasks used for policy evaluation.}
  \label{fig:robot_tasks}
\end{figure}

\subsection{Robot Setup and Tasks}
\label{app:robot_setup_tasks}

All real-robot experiments are conducted on an AgileX Piper robot arm. We collect
demonstrations through leader-follower teleoperation. The setup includes one
fixed third-person primary camera and one wrist-mounted camera. During policy
execution, the policy takes the task instruction, RGB observations from both cameras, and the current robot proprioception as input. The
policy outputs a 6-DoF end-effector pose command together with a continuous gripper
command.

We evaluate four manipulation tasks, as illustrated in
Fig.~\ref{fig:robot_tasks}. In \textit{Pick up Spoon}, the robot grasps a spoon
and places it onto a plate. In \textit{Stack Blocks}, the robot first stacks the
blue block on top of the green block and then stacks the red block on top of the
blue block. In \textit{Fold Towel}, the robot folds a towel twice to form a
square-like shape. In \textit{Open Drawer}, the robot opens a drawer, places a
purple cylinder inside, and then closes the drawer.

Each policy is evaluated with 20 independent real-robot trials per task. Object
positions are randomized before each trial. A trial is counted as successful only
if the full task sequence is completed and the final task state satisfies the
task-specific success criterion.

%% file: app/B.tex
\section{Evaluation Metrics}
\label{app:evaluation_metrics}
We evaluate each value model on held-out trajectories that are disjoint from the
annotated trajectories used for training. For each trajectory, we first evaluate the
value model on all 5 Hz frames and then linearly interpolate the predicted values
back to the raw-frame timeline. This produces a dense value trace \(v_i(t)\),
where \(t\) denotes the raw frame index.
We group held-out trajectories into three categories: clean successful
trajectories, suboptimal successful trajectories with retry behavior, referred to as retry trajectories below, and failed
trajectories. Global metrics evaluate whether the value model captures overall
task progress and final task outcome, while local retry metrics evaluate whether
the value trace exhibits the desired drop-and-rebound pattern around annotated
retry events.

\subsection{Global Metrics}
\label{app:global_metrics}

\paragraph{Value-Order Correlation (VOC).}
VOC measures whether the predicted value increases with task progress on clean
successful demonstrations. For each clean successful held-out trajectory, we compute the Spearman rank correlation between the 5 Hz frame index and the predicted value. We then report the average correlation over all clean successful trajectories. Recovery and failure trajectories are not included in VOC.

\paragraph{Success/Failure (S/F)  Detection.}
S/F Detection measures whether the terminal value correctly predicts the final
trajectory outcome. This metric is computed over all held-out successful and
failed trajectories, where successful trajectories include both clean successes
and retry trajectories. For each trajectory, we use the
value score at the last evaluated frame as the trajectory-level score. We use a fixed threshold of $0.9$ to distinguish successful and failed trajectories. The reported S/F Detection score is the
classification accuracy against the ground-truth trajectory outcome. All value scores are normalized to the same range before computing this metric.

\subsection{Local Retry Metrics}
\label{app:local_retry_metrics}
Local retry metrics evaluate whether a value model detects local mistakes and
recovery around annotated retry events. These metrics are computed only on
held-out recovery trajectories. We use a fixed evaluation radius \(K=30\) on the
raw-frame timeline. This radius is used both for computing local value drops and
for defining retry-centered evaluation windows. It is different from the
pre-retry, retry-centered, and post-recovery sampling windows used for training
in Sec.~\ref{sec:pair-window}; here it is used only to evaluate whether the
predicted value drops near the annotated retry keypoint and whether the retry
region forms a local value minimum.

Let \(\mathcal{R}_i\) denote the set of retry keypoints in trajectory \(i\), and
let \(v_i(t)\) be the interpolated value at raw frame \(t\). We compute a local
drop score at each raw frame by comparing the current value with the maximum
value within the preceding \(K\)-frame window:
\[
    D_i(t)
    =
    \max_{u\in[\max(0,t-K),t]} v_i(u) - v_i(t).
\]
A large \(D_i(t)\) indicates that the predicted value has recently dropped from
a nearby local maximum. For a retry keypoint \(r\), we use
\([r-K,r+K]\) as its retry-centered evaluation window.

\paragraph{Drop AUC.}
Drop AUC measures whether value drops are concentrated near annotated retry
events. For each retry keypoint, we use the maximum drop score within its
retry-centered window as the positive-window score. Negative
windows are sampled from regions at least \(2K\) raw frames away from any retry
keypoint, and their scores are computed in the same way. Let
\(\{(s_j,y_j)\}_{j=1}^{N}\) denote the resulting window-level examples, where
\(s_j\) is the maximum drop score in the window and \(y_j\in\{0,1\}\) indicates
whether the window is retry-centered. We sort windows by \(s_j\) in descending
order and compute the precision-recall AUC:
\[
    \mathrm{Drop~AUC}
    =
    \sum_{n=1}^{N}
    \left(R_n - R_{n-1}\right) P_n ,
\]
where \(P_n\) and \(R_n\) are the precision and recall after taking the top
\(n\) windows. A higher Drop AUC indicates that large value drops are more
localized around annotated retry events.

\paragraph{Drop Probability.}
Drop Probability measures how often annotated retries are accompanied by a
salient value drop. For each trajectory, we define a trajectory-specific
threshold
\[
    \eta_i
    =
    \max\left(
    \operatorname{Quantile}_{0.9}\{D_i(t):0\le t\le T_i\},
    0.01
    \right).
\]
We then report the fraction of retry keypoints whose retry-centered window
contains at least one drop score greater than \(\eta_i\). Unlike Drop AUC, this
metric does not use negative windows; it directly measures how consistently the
model produces a detectable value drop around annotated retry events.

\paragraph{Pre \(>\) Retry and Post \(>\) Retry.}
These metrics test whether the annotated retry keypoint corresponds to a local
low-value state. For each retry keypoint \(r\), we use the same evaluation
radius \(K\) defined above and compute the average values before and after the
retry keypoint:
\[
    v_i^{\mathrm{pre}}(r)
    =
    \frac{1}{K}
    \sum_{t=r-K}^{r-1} v_i(t),
    \qquad
    v_i^{\mathrm{post}}(r)
    =
    \frac{1}{K}
    \sum_{t=r+1}^{r+K} v_i(t).
\]
Boundary cases where the full pre- or post-window is not available are skipped.
\textbf{Pre \(>\) Retry} reports the fraction of retry events satisfying
\(v_i^{\mathrm{pre}}(r)>v_i(r)\), while \textbf{Post \(>\) Retry} reports the
fraction satisfying \(v_i^{\mathrm{post}}(r)>v_i(r)\). These metrics check
whether the value decreases toward the retry keypoint and recovers after
corrective behavior.

%% file: app/C.tex
\section{Retry Annotation Analysis}
\label{app:retry_annotation_details}

\subsection{Retry Keypoint Definition}
A retry keypoint marks the start of correction after a local execution error,
rather than the mistake onset or the completion of recovery. In our setting,
retry events mainly capture localized but consequential execution errors that
require immediate correction, such as misaligned grasps, missed grasps, or
inaccurate placements. 

We do not treat all execution errors as retry events. Instead, we only
keep cases with a clear localized correction signal. We exclude long off-task
error segments, such as grasping the wrong object and later completing the
task, collisions with the environment, failed corrections that do not improve
the state, and irreversible failures such as objects falling outside the
reachable workspace. This keeps retry supervision tied to recoverable local
execution errors rather than broad trajectory-level failures.

\subsection{Annotation Cost and Consistency}
\label{app:annotation_cost_consistency}

We further evaluate the annotation cost and temporal consistency of retry-related
labels. 
For annotation cost, three annotators independently label the value-model training
trajectories for each task. We report the average annotation time per trajectory and its variation across
annotators in Table~\ref{tab:annotation_cost}. This
measures the practical cost of obtaining sparse retry supervision. Table~\ref{tab:annotation_cost} shows that our sparse retry annotation can be
obtained efficiently: even for the slowest annotator, labeling all 30
trajectories for a task takes no more than 30 minutes, corresponding to less
than one minute per trajectory on average.

\begin{table*}[t]
\centering
\small
\caption{Annotation cost on the value-model training datasets. We report the total annotation time for each of three annotators
in minutes and the per-trajectory time across
annotators in seconds.}
\vspace{0.2cm}
\label{tab:annotation_cost}
\begin{tabular}{lcccccc}
\toprule
Task &
\# Retry/Succ./Fail Traj. &
\# Retry Keypts. &
Ann. 1 &
Ann. 2 &
Ann. 3 &
Sec./Traj. \\
\midrule
Pick up Spoon & 20/5/5 & 20& 10 min& 15 min& 8 min & $22.0 \pm 7.2$ \\
Stack Blocks  & 18/7/5 & 32  & 23 min& 26 min& 19 min& $45.3 \pm 7.0$ \\
Fold Towel    & 22/5/3 & 23  & 30 min& 27 min& 28 min& $56.7 \pm 3.1$ \\
Open Drawer   & 20/5/5 & 29  & 21 min& 25 min& 24 min& $46.7 \pm 4.2$ \\
\midrule
Average       & --     & 26.0 & 21.0 min& 23.3 min& 19.8 min& $42.7 \pm 14.1$ \\
\bottomrule
\end{tabular}
\end{table*}

We next evaluate the temporal consistency of retry-related annotations. 
We measure consistency from both inter-annotator and intra-annotator
perspectives. For inter-annotator consistency, three annotators independently
label the same retry events. For intra-annotator consistency, the same annotator
labels the same events three times. For each labeled timestep, let
$t_1,t_2,t_3$ denote the three labeled timestamps, either from three annotators
or three repeated passes. We compute the mean pairwise difference as
\[
    \mathrm{error}
    =
    \frac{
    |t_1 - t_2| + |t_1 - t_3| + |t_2 - t_3|
    }{3}.
\]
We report mean absolute timing errors averaged over retry events in
Table~\ref{tab:annotation_consistency}. All errors are reported in seconds.

Table~\ref{tab:annotation_consistency} shows that retry keypoints are temporally
consistent across annotators. The average inter-annotator error for retry
keypoints is only 0.19 seconds, which is less than one frame under our 5 Hz
annotation rate and is acceptable for value-model training. In contrast, the
auxiliary pre-retry and post-recovery timestamps are substantially less
consistent, reflecting the ambiguity of when degradation begins or recovery is
fully completed. This supports our design choice of using only the retry
keypoint as supervision and applying a soft temporal window around it, rather
than directly supervising the less reliable pre/post timestamps.

\begin{table*}[t]
\centering
\caption{Temporal consistency of retry-related annotations. Inter-annotator errors are computed across three annotators, and
intra-annotator errors are computed across three repeated passes by the same
annotator. All values are in seconds.}
\vspace{0.2cm}
\label{tab:annotation_consistency}
\begin{tabular}{lcccccc}
\toprule
Task &
\multicolumn{3}{c}{Inter-annotator error} &
\multicolumn{3}{c}{Intra-annotator error} \\
\cmidrule(lr){2-4}
\cmidrule(lr){5-7}
& Pre & Retry & Post & Pre & Retry & Post \\
\midrule
Pick up Spoon & 0.42 & 0.17 & 0.46 & 0.24 & 0.09 & 0.27 \\
Stack Blocks  & 0.55 & 0.14 & 0.72 & 0.16 & 0.09 & 0.20 \\
Fold Towel    & 0.64 & 0.27 & 0.81 & 0.30 & 0.14 & 0.42 \\
Open Drawer   & 0.43 & 0.19 & 0.63 & 0.21 & 0.16 & 0.31 \\
\midrule
Average       & 0.51 & 0.19 & 0.66 & 0.22 & 0.12 & 0.29 \\
\bottomrule
\end{tabular}
\end{table*}

%% file: app/D.tex
\section{Additional Experimental Results}
\subsection{Ablation Visualizations}
\begin{figure}[t]
  \centering
  \begin{subfigure}[t]{0.47\linewidth}
    \centering
    \includegraphics[width=\linewidth]{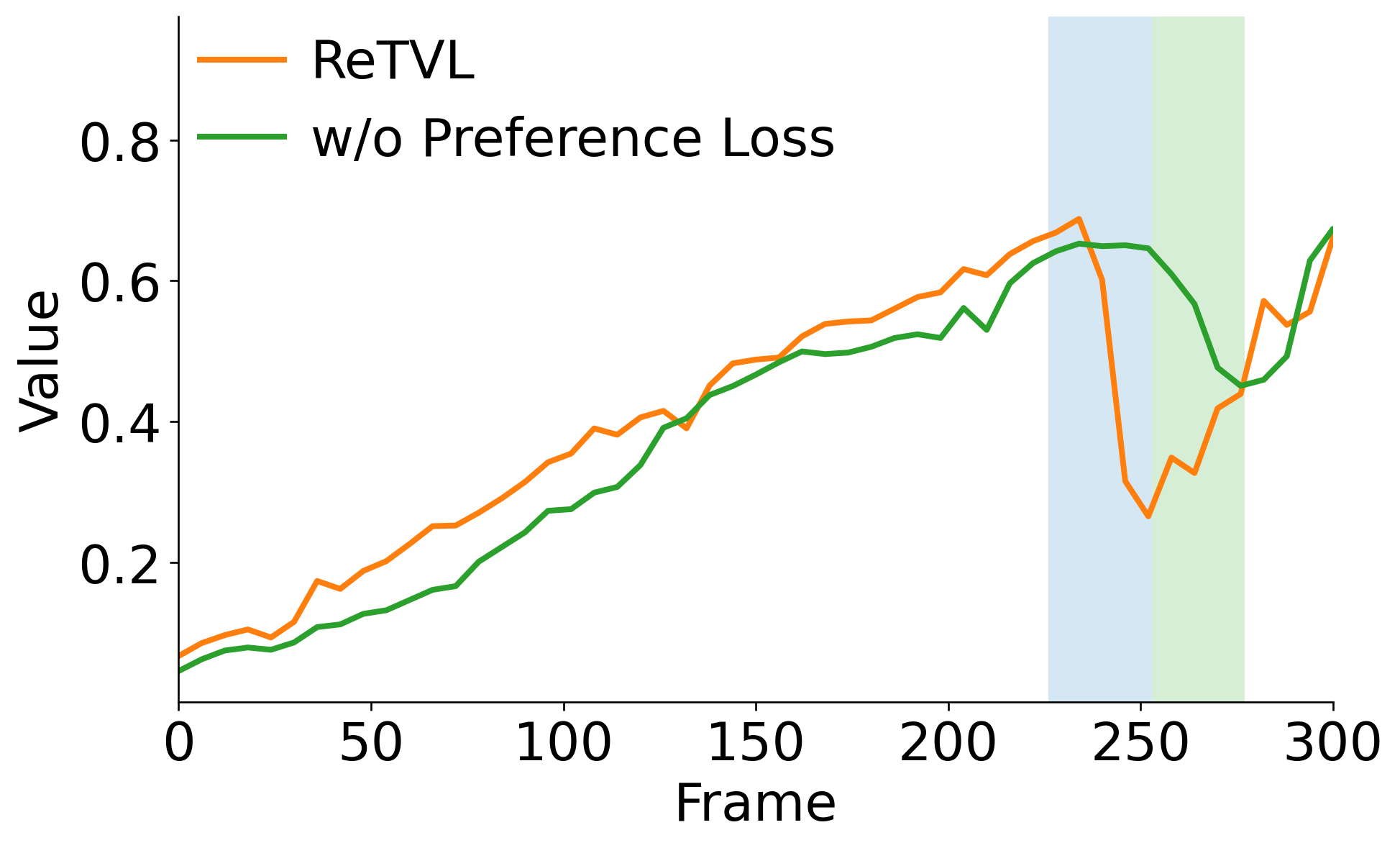}
  \end{subfigure}
  \hfill
  \begin{subfigure}[t]{0.47\linewidth}
    \centering
    \includegraphics[width=\linewidth]{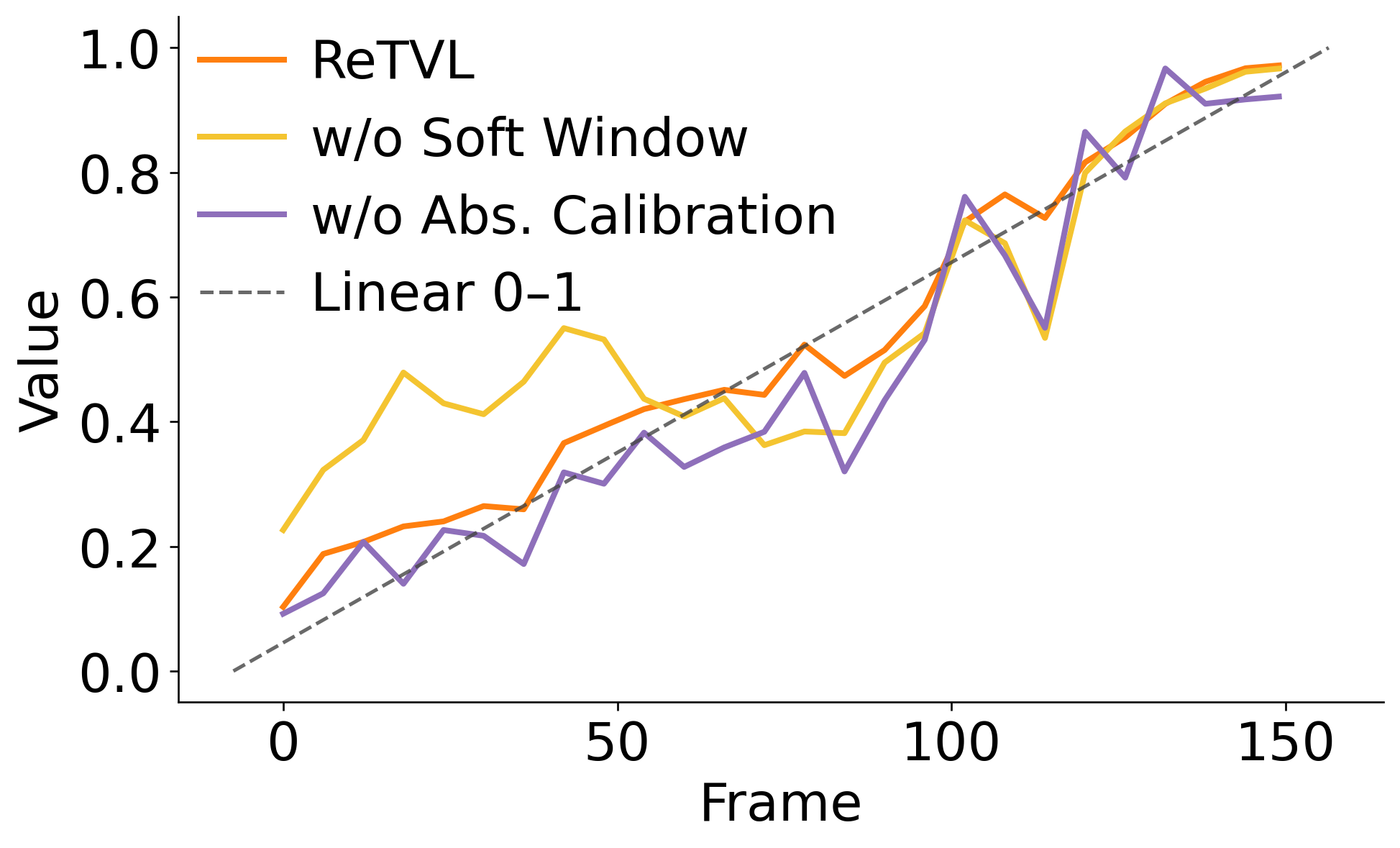}
  \end{subfigure}

  \caption{
   Representative value-curve visualizations for ablation variants on held-out trajectories.
    }
  \label{fig:ablation}
\end{figure}

We further provide representative value-curve visualizations to complement the
ablation results in Sec.~4.3. 
Figure~\ref{fig:ablation} shows that the drop-regression baseline without
preference loss does not reliably align value drops with annotated retry events.
Although this variant is trained with manually injected drops around retry
windows, its predicted value curve often exhibits misplaced or weak drops on
held-out trajectories. This suggests that directly shaping the absolute progress
target does not necessarily teach the model to recognize the visual meaning of
local execution errors. Instead, the model may fit the injected numeric pattern
in the training data without learning robust visual cues for mistake states.

The visualizations also show that removing soft-window weighting or absolute
calibration leads to less stable global progress estimates. Without soft-window
weighting, retry-related supervision is applied uniformly across approximate
retry neighborhoods, which can create sharper conflicts with the absolute
progress target near window boundaries. As a result, the value curve becomes
more oscillatory even on successful trajectories. Without absolute calibration,
the model relies mainly on local retry-induced preferences and loses a stable
global progress anchor, which similarly increases value fluctuations and weakens
the monotonic progress structure.


\subsection{Analysis of Downstream Weights}
\label{app:weight_analysis}

We further report the chunk-weighting results of ReTVL and the baselines on the test set. This analysis complements the retry-centered value-shape
metrics in Table~\ref{tab:value_eval}, as it directly evaluates whether the
resulting behavior-cloning weights preserve useful chunks and suppress harmful
chunks. This analysis is less tied to the training loss and therefore provides a more direct view of how the learned values affect downstream policy learning.

We report four metrics in Table~\ref{tab:weight_analysis}. \emph{Success Weight}
measures the average training weight assigned to chunks from successful
trajectories. \emph{Post-retry Weight} measures the average weight assigned to
manually annotated recovery chunks after retry. \emph{Success Deletion} measures
the fraction of successful chunks that are incorrectly assigned zero weight,
while \emph{Strict-bad Retention} measures the fraction of manually annotated
harmful chunks that are incorrectly retained with positive weight. Lower Success
Deletion and Strict-bad Retention indicate better filtering behavior.

\begin{table}[t]
\centering
\setlength{\tabcolsep}{5pt}
\caption{
Chunk-weighting analysis averaged over four real-robot tasks.
}
\label{tab:weight_analysis}
\begin{tabular}{lcccc}
\toprule
Method
& Success W. $\uparrow$
& Post-retry W. $\uparrow$
& Success Del. $\downarrow$
& Strict-bad Ret. $\downarrow$ \\
\midrule
TOPReward
& 0.539
& 0.498
& 0.330
& 0.695 \\
Robometer
& \textbf{0.845}
& 0.432
& \textbf{0.085}
& 0.896 \\
RECAP-Value
& 0.716
& 0.438
& 0.150
& 0.743 \\
ReTVL
& 0.769
& \textbf{0.636}
& 0.146
& \textbf{0.206} \\
\bottomrule
\end{tabular}
\end{table}

Table~\ref{tab:weight_analysis} shows that ReTVL achieves the most favorable
trade-off between preserving useful chunks and suppressing harmful ones.
Robometer assigns the highest weights to successful chunks and has the lowest
Success Deletion rate, but it also retains most strict-bad chunks, with a
Strict-bad Retention rate of 0.896. In contrast, ReTVL assigns the highest
weights to post-retry recovery chunks while reducing Strict-bad Retention to
0.206, substantially lower than all baselines. These results support the central
role of retry-supervised value learning: ReTVL does not merely suppress all
retry-related data, but learns a more selective weighting signal that preserves
useful recovery behavior and downweights harmful mistake segments.